\documentclass[11pt]{article}

\usepackage[preprint]{acl}

\usepackage{times}
\usepackage{latexsym}

\usepackage[T1]{fontenc}

\usepackage[utf8]{inputenc}

\usepackage{microtype}
\usepackage{tabularx}
\usepackage{multirow}

\usepackage{hyperref}

\usepackage{inconsolata}

\usepackage{graphicx}
\usepackage{xcolor}

\usepackage{mdframed}

\usepackage{amsmath}
\usepackage{amssymb}
\usepackage{enumitem}

\title{When Your LLM Reaches End-of-Life: A Framework for Confident Model Migration in Production Systems}

\author{\textbf{Emma Casey, David Roberts, David Sim, Ian Beaver} \\
         Verint Systems Inc, Melville, NY, USA \\
         \texttt{\{first.last\}@verint.com}}

\begin{document}
\maketitle
\begin{abstract}
  We present a framework for migrating production Large Language Model (LLM) based systems when the underlying model reaches end-of-life or requires replacement. The key contribution is a Bayesian statistical approach that calibrates automated evaluation metrics against human judgments, enabling confident model comparison even with limited manual evaluation data. We demonstrate this framework on a commercial question-answering system serving 5.3M monthly interactions across six global regions; evaluating correctness, refusal behavior, and stylistic adherence to successfully identify suitable replacement models. The framework is broadly applicable to any enterprise deploying LLM-based products, providing a principled, reproducible methodology for model migration that balances quality assurance with evaluation efficiency — a capability increasingly essential as the LLM ecosystem continues to evolve rapidly and organizations manage portfolios of AI-powered services across multiple models, regions, and use cases.
\end{abstract}

\section{Introduction}

Commercial products that make extensive use of Large Language Models (LLMs) are increasingly pervasive in business. Our company currently offers over a dozen enterprise products using LLMs for various features.  Often the LLMs used are proprietary or too large to self-host and are instead hosted by third parties such as Azure, Amazon Web Services, or Google Cloud Platform.  A challenge unique to supporting these products from traditional software is what to do when the LLM with which a product has been developed, tested and field-hardened reaches end-of-life and a replacement needs to be selected. Other reasons driving LLM migration include deploying the product in a new geographical region or platform where the original LLM is unavailable, unsatisfactory performance in a new language, cost optimization for the product, or improving latency and responsiveness.  This proprietary LLM deprecation and migration cycle has been happening roughly every 12 months and impacts every product built on the deprecated model, meaning a target model must be identified and migrated to for all impacted products before the LLM is removed from the hosting platform.

Robust task-specific evaluation seems essential, but an inefficient and time-consuming manual process is prohibitively expensive given the short migration windows, frequency of migrations, and breadth of impacted products. Evaluations can be complex as they are influenced by multiple aspects of the quality of the product's output given the new model, as well as by operational restrictions on the models such as cost and regional availability. Various public test sets and tools may be available, but they may not always align to the business-derived goals of the system. The process of evaluation is further complicated by the coupling of model and prompt, and the need, ideally, to avoid evaluating a potential model with a suboptimal prompt. In some respects however, the task is simplified compared to the initial development and testing of the product because we have the initial model as a benchmark for acceptable quality.

In this paper we present a real world case study on a commercial question-answering system demonstrating the framework we developed to speed and reduce cost of LLM migrations in our products. This system averages 5.3M chats per month across 6 global regions, supports multiple languages, and has been in commercial usage for over two years primarily supporting contact center agents during customer interactions for dozens of enterprise companies.  Maintaining output quality for current enterprise customers, ensuring adequate speed across deployed geographic regions, and not substantially increasing LLM cost are the primary constraints of model choice. We offer the following contributions:

\begin{enumerate}[noitemsep]
  \item{We present our enterprise tested framework developed for selecting a new model and adapting the current prompt to work with it. Aspects of the precise method shown are tailored to the case study system, but the principles and techniques underpinning it are transferable to other LLM-based systems.}
  \item{We reflect on the impact and effectiveness of different methods for adapting an existing prompt to a new model.}
\end{enumerate}

\section{Related Work}

Since its introduction in \citet{lewis2020retrieval}, Retrieval-Augmented Generation (RAG) has become a dominant paradigm for building question-answering systems \citep{gupta2024comprehensive,rakin2024leveraging}. A typical RAG system combines a dense retriever, which selects relevant context from a knowledge store, with an LLM-based answer generation component \citep{sharma2025retrieval}. The popularity of RAG has driven the development of numerous evaluation tools targeting both retrieval and generation components.

The usefulness of a question-answering system depends on multiple characteristics, which vary depending on the application \citep{gehrmann2023repairing}. Test sets such as HotpotQA \citep{yang-etal-2018-hotpotqa} and SQuAD \citep{rajpurkar-etal-2016-squad} provide structured data that enable evaluation of the retrieval and answer-generation components either independently or jointly.

Early evaluations compared generated and gold-standard answers using word-overlap metrics such as ROUGE and BLEU. However, the open-ended nature of question answering limits the effectiveness of these metrics \citep{krishna2021hurdles}. More sophisticated methods now incorporate semantic matching \citep{conf/iclr/ZhangKWWA20,wang2023chatgpt}, and many of these techniques are part of widely used toolkits such as RAGAS \citep{es-etal-2024-ragas}.

When migrating a RAG system between LLMs, care must be taken to ensure that prompts effective for one model continue to perform well on another \citep{chen2024mapoboostinglargelanguage,wang2025promptbridgecrossmodelprompttransfer}. This motivates model-specific prompt tuning. Model providers typically publish prompting guidelines tailored to their models, and automated prompt optimizers—such as MIPROv2 \citep{opsahlong2024optimizinginstructionsdemonstrationsmultistage} and GEPA \citep{agrawal2025gepareflectivepromptevolution}—could, in principle, adapt prompts to new LLMs just as they optimize prompts for new tasks.

Despite this, relatively little academic work has focused specifically on the manual or automatic adaptation of existing prompts to new models. While \citet{jahani2026promptadaptationdynamiccomplement} suggest that prompt adaptation can be a primary driver of improvements when switching models, they report that automatic prompt rewriting is not yet reliably effective.

\section{Case Study System}

The system requiring migration is a question-answering service shared across several customer-support products. Given a user query and a ranked set of source texts, it returns a natural-language answer with citations. At its core is a prompt instructing an LLM to generate answers strictly from the provided sources and to follow specific formatting guidelines. The model is required to return a structured XML response, including a flag indicating when it cannot answer the question, enabling a clean ``I don't know'' (IDK) output.

This system has been deployed extensively in production, and its output quality is considered sufficient for current business use. However, as its underlying LLM—Claude 3 Haiku—approaches end-of-life, we must migrate to an alternative model without degrading performance. The pool of candidate models is restricted to those that have passed internal vetting for training bias, privacy and compliance risk, licensing constraints, operating cost limits, and availability across all hosting regions used by our customer base.

Our quality requirements are multifaceted. First, we require an acceptable rate of correct answers. Unlike many standard QA evaluations that focus primarily on accuracy, we consider an IDK response preferable to a confident but incorrect answer. However, an excessively high IDK rate on answerable questions would also be unacceptable. While the potential harm of incorrect answers is also relevant in practice, we use the rate of wrong answers as a proxy for this during evaluation.

Answers must also adhere to a style appropriate for business applications: responses must consist of complete sentences, be reasonably concise, and adopt a tone consistent with a human expert. In particular, they should avoid stock phrases such as ``according to my sources'' or ``according to the information provided.''

Adapting the prompt to the new model is possible and may affect any of these quality dimensions, positively or negatively. However, our goal is to minimize the time spent experimenting with alternative prompts in order to reduce both migration cost and overall timeline.

\subsection{Evaluation Tools}

There are many test sets that provide combinations of questions, answers, and ground truth, but their reliability varies. Public datasets are typically large but often poorly aligned with our business domains, while limited annotation capacity means that system-specific test sets are relatively small.

Automated evaluation is desirable—particularly because LLM migrations are a recurring challenge for enterprise systems—but current tools for assessing correctness are imperfect and may be misaligned with our requirements. For example, most automated metrics classify IDK responses as incorrect, whereas we treat them as a distinct (though still undesirable) category. In addition, the limitations of gold-standard answers mean that relying solely on the ground truth without reference to the full source context may lead to valid answers being incorrectly penalised.


\section{Framework for Model Selection}

\begin{enumerate}
  \setlength{\itemsep}{0pt}
  \item Select candidate models that have passed any internal vetting process and cost limitations.
  \item Eliminate any models which struggle to generate output that follows any required output schema such as XML or JSON.
  \item For each test set count the number of IDK responses, and evaluate the correctness of each candidate model relative to the baseline using the methodology described below.
  \item Eliminate any models which display a significant increase in IDK rate or response time, or a measurable decrease in correctness relative to the baseline on any test set.
  \item Check the output of the remaining models for style violations. If necessary, we attempt to fix style issues by improving the prompt, then re-evaluate correctness and style for the new prompts.
  \item Choose a subset of the remaining models that cover all required combinations of region and supported modality. To cover all regional requirements, more than one LLM may need to be supported. If multiple options are available, a final decision can be based on optimal price and quality.
\end{enumerate}

\begin{table*}[th]
  \centering
  \small
  \begin{tabular}{|l|l|c|c|c|c|}
    \hline
    \textbf{Test-set}       & \textbf{Metric}    & \textbf{TPR Mean} & \textbf{TPR 90\% CI}    & \textbf{FPR Mean} & \textbf{FPR 90\% CI}    \\
    \hline
    \multirow{5}{*}{basic}  & ragas\_correctness & 0.762             & [0.648, 0.861]          & 0.688             & [0.489, 0.858]          \\
                            & llm\_correctness   & 0.857             & [0.761, 0.934]          & 0.500             & [0.311, 0.689]          \\
                            & new\_correctness   & \textbf{0.905}    & \textbf{[0.822, 0.966]} & \textbf{0.375}    & \textbf{[0.191, 0.577]} \\
                            & faithfulness       & 0.881             & [0.790, 0.951]          & 0.750             & [0.560, 0.903]          \\
                            & relevance          & 0.810             & [0.703, 0.899]          & 0.812             & [0.637, 0.943]          \\
    \hline
    \multirow{5}{*}{hotpot} & ragas\_correctness & 0.750             & [0.654, 0.836]          & 0.200             & [0.041, 0.429]          \\
                            & llm\_correctness   & 0.850             & [0.769, 0.918]          & 0.300             & [0.098, 0.550]          \\
                            & new\_correctness   & \textbf{0.883}    & \textbf{[0.809, 0.943]} & \textbf{0.100}    & \textbf{[0.006, 0.283]} \\
                            & faithfulness       & 0.800             & [0.710, 0.878]          & 0.900             & [0.717, 0.994]          \\
                            & relevance          & 0.667             & [0.564, 0.763]          & 0.400             & [0.169, 0.655]          \\
    \hline
  \end{tabular}
  \caption{True positive rates (TPR) and false positive rates (FPR) of QA metrics compared to human grading}
  \label{tab:metric-rates}
\end{table*}

\subsection{Correctness Comparison}
Since the automatic evaluation metrics aren't perfectly accurate and data is limited, a reported difference in numbers of correct responses might not reflect a real difference in correctness rates. To account for this we use a Bayesian analysis to compare models. We estimate metric accuracy by manually evaluating a subset of examples, before using the metric to evaluate the model. Using this we can describe the difference in correctness between models in terms of confidence intervals which take account of metric accuracy as well as the limited sample sizes in both the manual and automated parts of the evaluation.

Mathematical details are in Appendix~\ref{appendix:bayesian-math}. The manual evaluation only needs to be done once for a given test set, and can then be reused for other metrics and models.

\subsection{Style Checking}

For each experimental run we check the style of the output. In particular, we use a simple substring match to check for a number of unwanted features that we've observed in previous versions of the system: referring to "the sources", "the knowledge" or saying "according to". We report these individually and also report "bad style" in general for any response where one or more of these is seen. We also report the word count for the answers, to check for any tendency towards unexpectedly verbose or terse responses.

\section{Experimental Setup}

Our baseline model was Claude 3 Haiku. The models passing the first step of our framework were Gemma 3 (27B), Anthropic Claude 4.5 Haiku and 3.5 Sonnet, AWS Nova Micro, Lite, Lite 2 and Pro; OpenAI GPT-OSS 20B and 120B, Qwen3-32B and 235B. For Qwen3-32B we evaluated the system with and without the model's reasoning mode enabled. Many other potential models were eliminated by our internal vetting process which aims to ensure the hosted LLM does not pose a significant business risk due to bias, data residency or licensing issues for the enterprise application, or they required excessive dedicated hosting costs to maintain data residency in non-USA regions.

\subsection{Test Sets}
We evaluated against 200 examples from the HotpotQA test set, 200 from the SQuAD test set, and 51 samples from an internal test set representing natural language questions about a set of customer-service-style documents.

From the output of the full set of models over the three test sets, we randomly selected 66 HotpotQA examples, 55 from the internal test set and 25 from SQuAD to evaluate manually to assess the metrics. A group of three assessors judged these collaboratively and without access to the metrics' judgments on them. In order to arrive at consistent evaluations of a number of subtle failure modes, the assessors had to agree on guidelines for how to handle marginal cases as they went. A sample of challenging cases are shown in Appendix~\ref{sec:appendix}.

Although the numbers of examples here are relatively small and will lead to correspondingly high uncertainty in our results, our Bayesian analysis will allow us to quantify this uncertainty and account for it in our decision making.

\subsection{Metrics} \label{sec:metrics}
The correctness metrics considered were the correctness, faithfulness and relevance metrics from the RAGAS test suite, and two LLM-based evaluations of our own: \textbf{llm\_correctness,} which uses an LLM to match an answer against the ground truth and
\textbf{new\_correctness} which evaluates answers against the full source context rather than the ground truth alone, better capturing cases where ground truths are incomplete or incorrect. Prompts are given in Appendix~\ref{appendix:prompts} and were run against Claude 4 Sonnet.

\section{Results} \label{sec:results}

The second step of our framework eliminated both OpenAI GPT-OSS models as they were unable to consistently generate output following the XML schema required by the product. Qwen3-32B suffered from a similar problem, but this was quickly resolved by adding examples of correctly formatted output to the prompt, so we applied this modification to the prompt used for this model in the subsequent experiments.

\subsection{Correctness Metric Evaluation}
Using confusion matrices constructed for each metric based on our manual evaluation of answers from each test set, we estimated posterior distributions for the true-positive and false-positive rates for the various metrics as given in Table~\ref{tab:metric-rates}.

After our initial evaluation, we opted to discard the examples from SQuAD as we had found insufficient true-negatives to effectively calibrate the metrics. This left us with our internal test set, labeled ``basic'', and HotpotQA.  Based on these results, new\_correctness was the metric most closely aligned with actual correctness on both test sets, and is used for our remaining comparisons.

\subsection{Model Correctness Comparisons}

Table~\ref{tab:model-comparison} shows how the remaining models compare to Claude 3 Haiku by estimated ``true'' correctness rates of the new\_correctness metric and IDK rates.

\begin{table}[t]
  \centering
  \footnotesize
  \setlength{\tabcolsep}{2.5pt}
  \begin{tabular}{|l|l|r|r|c|}
    \hline
    \textbf{Set} & \textbf{Model}    & \textbf{IDK} & \textbf{Diff} & \textbf{90\% CI}   \\
    \hline
    \multirow{11}{*}{\rotatebox{90}{basic}}
                 & Claude 3 Haiku    & 11.7         & ---           & ---                \\
                 & Claude 3.5 Sonnet & 3.19         & 9.89          & [4.59, 15.8]       \\
                 & Claude 4.5 Haiku  & 4.26         & 12.8          & [6.46, 19.9]       \\
                 & Nova Micro        & 11.7         & -9.33         & [$-15.9$, $-3.57$] \\
                 & Nova Lite         & 4.26         & 0.74          & [$-4.75$, 6.36]    \\
                 & Nova 2 Lite       & 2.13         & 4.85          & [0.173, 10.1]      \\
                 & Nova Pro          & 5.32         & 5.64          & [0.688, 11.2]      \\
                 & Gemma 3 27B       & 5.32         & 0.643         & [$-4.75$, 5.97]    \\
                 & Qwen3-32B (r)     & 4.26         & 5.62          & [0.23, 11.62]      \\
                 & Qwen3-32B         & 3.19         & 9.79          & [4.16, 16.17]      \\
                 & Qwen3-235B        & 1.06         & 9.1           & [4.16, 14.7]       \\
    \hline
    \multirow{11}{*}{\rotatebox{90}{hotpot}}
                 & Claude 3 Haiku    & 5.5          & ---           & ---                \\
                 & Claude 3.5 Sonnet & 7.5          & 7.19          & [2.96, 11.7]       \\
                 & Claude 4.5 Haiku  & 8.0          & 8.15          & [3.91, 12.7]       \\
                 & Nova Micro        & 4.0          & -1.28         & [$-5.6$, 2.97]     \\
                 & Nova Lite         & 10.5         & 2.38          & [$-1.88$, 6.78]    \\
                 & Nova 2 Lite       & 4.0          & 4.75          & [0.692, 9.04]      \\
                 & Nova Pro          & 4.0          & 6.83          & [2.46, 11.4]       \\
                 & Gemma 3 27B       & 2.5          & 3.38          & [$-0.68$, 7.61]    \\
                 & Qwen3-32B (r)     & 2.0          & 5.54          & [1.97, 9.25]       \\
                 & Qwen3-32B         & 3.5          & 6.39          & [2.66, 10.29]      \\
                 & Qwen3-235B        & 4.5          & 5.59          & [1.73, 9.73]       \\
    \hline
  \end{tabular}
  \caption{Model comparison using new\_correctness against Claude 3 Haiku. (r) = reasoning mode, Diff = mean true correctness difference, CI = 90\% confidence interval.}
  \label{tab:model-comparison}
\end{table}

In step 4 of our framework, we eliminate Nova Micro, Nova Lite and Gemma 3 27B. The confidence intervals for the change in their correctness rates from Claude 3 Haiku include negative values, so we cannot confidently claim that they improve on the baseline question-answering behavior.

\subsection{Style Metrics and Response Time}

We evaluated style and performance metrics for the remaining candidate models (detailed results in Appendix~\ref{appendix:style-metrics}). In step 4 of our framework, Claude 3.5 Sonnet is eliminated due to slower response times (1.73s and 1.89s median response times on the hotpot and basic test sets respectively, compared to 1.19s and 1.06s for the baseline). Reasoning mode hurt the style performance of Qwen3-32B, increasing bad style occurrences from 2.1\% to 7.5\% on the basic test set, so we drop that variation in step 5 of our framework.

\subsection{Model Selection}

Table \ref{tab:full-comparison} shows the final five candidate models, their supported regions and their relative prices at the task, at time of our migration. Now at step 6 of our framework, we select Nova 2 Lite in situations where file processing is needed, and Qwen3-32B with reasoning disabled as the primary LLM to migrate this system to. Nova 2 was trained in over 200 languages and Qwen3-32B was trained in 119 languages and has demonstrated good machine translation performance in over 33 languages~\cite{zheng2025hunyuanmttechnicalreport}, both meeting our multilingual business needs. This choice offers a quality of service that will be equal to or better than the current system while offering full regional coverage at reduced financial cost and response time improvement.

\begin{table*}[t]
  \centering
  \scriptsize
  \setlength{\tabcolsep}{2pt}
  \begin{tabular}{|l|c|c|c|c|c|c|}
    \hline
                            & \textbf{Claude 3 Haiku} & \textbf{Claude 4.5 Haiku} & \textbf{Nova Pro} & \textbf{Nova 2 Lite} & \textbf{Qwen3-32B} & \textbf{Qwen3-235B} \\
                            & \textbf{(Baseline)}     &                           &                   &                      &                    &                     \\
    \hline
    Price Tier              & Middle                  & High                      & High              & Middle               & Low                & Middle              \\
    EMEA Region 1$^\dagger$ & \checkmark              & ---                       & \checkmark        & \checkmark           & \checkmark         & \checkmark          \\
    APAC Region 1$^\dagger$ & \checkmark              & \checkmark                & \checkmark        & ---                  & \checkmark         & ---                 \\
    AMER Region 2$^\dagger$ & \checkmark              & ---                       & ---               & ---                  & \checkmark         & ---                 \\
    APAC Region 2$^\dagger$ & \checkmark              & ---                       & ---               & ---                  & \checkmark         & ---                 \\
    File Processing         & \checkmark              & \checkmark                & \checkmark        & \checkmark           & ---                & ---                 \\
    Correctness             & Low                     & Best                      & Good              & Good                 & Good               & Good                \\
    \hline
  \end{tabular}
  \caption{Price, regional availability outside of USA, and quality of remaining models. $^\dagger$Includes self-hosting options. Correctness ranking based on mean difference estimates from Table \ref{tab:model-comparison}.}
  \label{tab:full-comparison}
\end{table*}

\begin{table*}[t]
  \centering
  \footnotesize
  \setlength{\tabcolsep}{4pt}
  \begin{tabular}{|l|c|c|c|c|c|c|}
    \hline
    \textbf{Testset}       & \multicolumn{3}{|c|}{Basic} & \multicolumn{3}{|c|}{HotpotQA}                                                                               \\
    \hline
    \textbf{Model}         & \textbf{\% IDK}             & \textbf{Diff}                  & \textbf{90\% CI}       & \textbf{\% IDK} & \textbf{Diff} & \textbf{90\% CI} \\
    \hline
    Baseline               & 2.1                         & ---                            & ---                    & 4.0             & ---           & ---              \\
    aws                    & 1.1                         & 2.53                           & [-1.60, 7.02]          & 2.5             & -3.29         & [-6.63, -0.13]   \\
    dspy - f1              & 7.5                         & \textbf{7.99}                  & \textbf{[3.03, 13.72]} & 7.5             & -0.45         & [-4.00, 3.12]    \\
    dspy - new correctness & 6.4                         & \textbf{6.58}                  & \textbf{[1.68, 12.16]} & 6.5             & -2.15         & [-5.89, 1.42]    \\
    manual - capitals      & 7.4                         & 2.64                           & [-2.04, 7.62]          & 5.0             & 1.69          & [-1.08, 4.56]    \\
    manual with evidence   & 6.4                         & 1.99                           & [-2.27, 6.45]          & 5.0             & -0.03         & [-2.67, 2.63]    \\
    \hline
  \end{tabular}
  \caption{Prompt adaptation comparisons. Diff = mean true correctness difference, CI = 90\% confidence interval.}
  \label{tab:prompt-comparison}
\end{table*}

\section{Experiments with Prompt Adaptation}
Having identified a set of candidate models that satisfy our operational and quality requirements, we next investigated whether prompt adaptation could further improve system performance.

We explored three strategies - for each one, we generated revised prompts and evaluated our candidate models with the same correctness and style metrics used in Section~\ref{sec:results}. For brevity, we will discuss the results for Nova 2 Lite.

The first strategy was manual adaptation based on the model provider's recommended practices. Following this, we produced two variant prompts, one with key instructions given in capitals, and one additionally requesting evidence chunks from the context. Our second strategy used the prompt-optimization tool \cite{Bai_Dewan_Imran_Wang_2025} within Amazon Bedrock's Prompt Management suite. We provided the baseline prompt and requested an optimized version targeted to Nova 2 Lite.

Finally, we applied MIPROv2 through DSPy - details of this process are given in Appendix~\ref{appendix:dspy}. Here we give results for prompts trained using new\_correctness metric and tokenwise f1 similarity on a subset of HotpotQA.
\subsection{Results of Prompt Adaptation}
Table \ref{tab:prompt-comparison} shows evaluation results for Nova 2 Lite using these adapted prompts compared to the Claude 3 Haiku-optimized baseline prompt.  Automatic and manual adaptation failed to produce prompts that are clearly better than the original, although the use of capital letters to emphasize key instructions shows some evidence of improvement and might justify further testing.

While the dspy models performed well against our basic test set they substantially increased IDK rates, making fewer errors by refusing to answer more often. This is better behaviour, but these improvements don't replicate in the HotpotQA test set where accuracy decreased. This suggests that while MIPROv2 might be helpful in adapting an existing prompt to a new LLM we need to do more work to be confident that any of the prompts was a general improvement across many contexts.

Based on this data, the baseline prompt generalizes surprisingly well across models. However, better testing or more extensive prompt-engineering might still find better alternatives.

\section{Future Work}
\begingroup\setlength{\parskip}{2pt}%
There are a number of ways in which we intend to build on the analysis that was used in this paper:

\paragraph{Apply the process to other services.} We plan to apply this framework on the migration of many other LLM-based products. To do this, we will need to find or create system appropriate data sets and metrics, but the core framework can apply directly to any system with a reasonably well-defined concept of correctness.
\paragraph{Improve the precision of our analysis for Question Answering.} Increasing the size and variety of data sets and developing improved automated metrics will allow us to apply the same evaluation process to smaller changes and enable faster migrations to new LLMs in the future.
\paragraph{Integrating our comparison framework into a continuous evaluation system} would allow monitoring of drift, assessment of vendor updates and efficient evaluation of candidate models, ensuring the framework remains effective as model ecosystems continue to evolve.
\endgroup

\section{Conclusions}

We presented a structured framework for migrating production LLMs in a RAG-based QA system. By decomposing the task into stages---candidate filtering, metric calibration, Bayesian correctness comparison, style and latency checks, and finally prompt-adaptation experiments---we reduced evaluation complexity and ensured decisions remained traceable and reproducible.

A key finding is that reliable model selection requires aligned evaluation metrics rather than blind reliance on standard QA benchmarks. By calibrating automated metrics against manually judged examples, we identified substantial variability in their accuracy, with only new\_correctness consistently matching human judgment. Incorporating these calibrated metrics into a Bayesian comparison procedure provided meaningful confidence intervals for correctness differences, enabling more defensible model choices.

Applied to our commercial QA system, the framework allowed us to eliminate unsuitable candidates and identify Nova 2 Lite and Qwen3-32B as robust replacements that satisfy correctness, refusal behavior, style, and operational constraints.

Experiments with manual, automated, and machine-learned prompt adaptation showed no statistically significant improvements over the baseline prompt, suggesting either strong prompt transferability or the need for more intensive model-specific tuning.

Overall, the framework offers an efficient and principled method for enterprise LLM migration, providing dependable quality assurance while limiting evaluation cost.

\section{Limitations}

Because SQuAD was dropped for having insufficient calibration data, the main analysis relied on two test sets. The analysis was also restricted to English language, and further experiments would be needed to check response quality in other languages.

Metric calibration and autoevaluation used relatively small test sets. This reduces the precision of the comparisons which may have led to some models being unnecessarily eliminated from consideration.

LLM‑based metrics used a single judge model, which may bias metric behavior. The ability to use ground truth answers will mitigate the impact in this context, but a mixture-of-models approach might be desirable in the future.

Style evaluation targeted specific issues that have been seen before. There's scope for more robust checks at this stage, but we expect that any new issues will be dealt with during our usual Quality Assurance process.

\bibliography{custom}

@inproceedings{rajpurkar-etal-2016-squad,
  title     = {{SQ}u{AD}: 100,000+ Questions for Machine Comprehension of Text},
  author    = {Rajpurkar, Pranav  and
               Zhang, Jian  and
               Lopyrev, Konstantin  and
               Liang, Percy},
  editor    = {Su, Jian  and
               Duh, Kevin  and
               Carreras, Xavier},
  booktitle = {Proceedings of the 2016 Conference on Empirical Methods in Natural Language Processing},
  month     = nov,
  year      = {2016},
  address   = {Austin, Texas},
  publisher = {Association for Computational Linguistics},
  url       = {https://aclanthology.org/D16-1264/},
  doi       = {10.18653/v1/D16-1264},
  pages     = {2383--2392}
}

@inproceedings{yang-etal-2018-hotpotqa,
  title     = {{H}otpot{QA}: A Dataset for Diverse, Explainable Multi-hop Question Answering},
  author    = {Yang, Zhilin  and
               Qi, Peng  and
               Zhang, Saizheng  and
               Bengio, Yoshua  and
               Cohen, William  and
               Salakhutdinov, Ruslan  and
               Manning, Christopher D.},
  editor    = {Riloff, Ellen  and
               Chiang, David  and
               Hockenmaier, Julia  and
               Tsujii, Jun{'}ichi},
  booktitle = {Proceedings of the 2018 Conference on Empirical Methods in Natural Language Processing},
  month     = oct # {-} # nov,
  year      = {2018},
  address   = {Brussels, Belgium},
  publisher = {Association for Computational Linguistics},
  url       = {https://aclanthology.org/D18-1259/},
  doi       = {10.18653/v1/D18-1259},
  pages     = {2369--2380}
}

@article{wang2023chatgpt,
  title   = {Is chatgpt a good nlg evaluator? a preliminary study},
  author  = {Wang, Jiaan and Liang, Yunlong and Meng, Fandong and Sun, Zengkui and Shi, Haoxiang and Li, Zhixu and Xu, Jinan and Qu, Jianfeng and Zhou, Jie},
  journal = {arXiv preprint arXiv:2303.04048},
  year    = {2023}
}

@inproceedings{conf/iclr/ZhangKWWA20,
  author    = {Zhang, Tianyi and Kishore, Varsha and Wu, Felix and Weinberger, Kilian Q. and Artzi, Yoav},
  booktitle = {ICLR},
  ee        = {https://openreview.net/forum?id=SkeHuCVFDr},
  publisher = {OpenReview.net},
  title     = {BERTScore: Evaluating Text Generation with BERT.},
  url       = {http://dblp.uni-trier.de/db/conf/iclr/iclr2020.html#ZhangKWWA20},
  year      = 2020
}

@article{gehrmann2023repairing,
  title   = {Repairing the cracked foundation: A survey of obstacles in evaluation practices for generated text},
  author  = {Gehrmann, Sebastian and Clark, Elizabeth and Sellam, Thibault},
  journal = {Journal of Artificial Intelligence Research},
  volume  = {77},
  pages   = {103--166},
  year    = {2023}
}

@inproceedings{es-etal-2024-ragas,
  title     = {{RAGA}s: Automated Evaluation of Retrieval Augmented Generation},
  author    = {Es, Shahul  and
               James, Jithin  and
               Espinosa Anke, Luis  and
               Schockaert, Steven},
  editor    = {Aletras, Nikolaos  and
               De Clercq, Orphee},
  booktitle = {Proceedings of the 18th Conference of the European Chapter of the Association for Computational Linguistics: System Demonstrations},
  month     = mar,
  year      = {2024},
  address   = {St. Julians, Malta},
  publisher = {Association for Computational Linguistics},
  url       = {https://aclanthology.org/2024.eacl-demo.16/},
  doi       = {10.18653/v1/2024.eacl-demo.16},
  pages     = {150--158}
}

@article{krishna2021hurdles,
  title   = {Hurdles to progress in long-form question answering},
  author  = {Krishna, Kalpesh and Roy, Aurko and Iyyer, Mohit},
  journal = {arXiv preprint arXiv:2103.06332},
  year    = {2021}
}

@article{lewis2020retrieval,
  title   = {Retrieval-augmented generation for knowledge-intensive nlp tasks},
  author  = {Lewis, Patrick and Perez, Ethan and Piktus, Aleksandra and Petroni, Fabio and Karpukhin, Vladimir and Goyal, Naman and K{\"u}ttler, Heinrich and Lewis, Mike and Yih, Wen-tau and Rockt{\"a}schel, Tim and others},
  journal = {Advances in neural information processing systems},
  volume  = {33},
  pages   = {9459--9474},
  year    = {2020}
}

@article{gupta2024comprehensive,
  title   = {A comprehensive survey of retrieval-augmented generation (rag): Evolution, current landscape and future directions},
  author  = {Gupta, Shailja and Ranjan, Rajesh and Singh, Surya Narayan},
  journal = {arXiv preprint arXiv:2410.12837},
  year    = {2024}
}

@article{rakin2024leveraging,
  title   = {Leveraging the domain adaptation of retrieval augmented generation models for question answering and reducing hallucination},
  author  = {Rakin, Salman and Shibly, Md AR and Hossain, Zahin M and Khan, Zeeshan and Akbar, Md Mostofa},
  journal = {arXiv preprint arXiv:2410.17783},
  year    = {2024}
}

@article{sharma2025retrieval,
  title   = {Retrieval-Augmented Generation: A Comprehensive Survey of Architectures, Enhancements, and Robustness Frontiers},
  author  = {Sharma, Chaitanya},
  journal = {arXiv preprint arXiv:2506.00054},
  year    = {2025}
}

@misc{opsahlong2024optimizinginstructionsdemonstrationsmultistage,
  title         = {Optimizing Instructions and Demonstrations for Multi-Stage Language Model Programs},
  author        = {Krista Opsahl-Ong and Michael J Ryan and Josh Purtell and David Broman and Christopher Potts and Matei Zaharia and Omar Khattab},
  year          = {2024},
  eprint        = {2406.11695},
  archiveprefix = {arXiv},
  primaryclass  = {cs.CL},
  url           = {https://arxiv.org/abs/2406.11695}
}

@misc{chen2024mapoboostinglargelanguage,
  title         = {MAPO: Boosting Large Language Model Performance with Model-Adaptive Prompt Optimization},
  author        = {Yuyan Chen and Zhihao Wen and Ge Fan and Zhengyu Chen and Wei Wu and Dayiheng Liu and Zhixu Li and Bang Liu and Yanghua Xiao},
  year          = {2024},
  eprint        = {2407.04118},
  archiveprefix = {arXiv},
  primaryclass  = {cs.CL},
  url           = {https://arxiv.org/abs/2407.04118}
}

@misc{wang2025promptbridgecrossmodelprompttransfer,
  title         = {PromptBridge: Cross-Model Prompt Transfer for Large Language Models},
  author        = {Yaxuan Wang and Quan Liu and Zhenting Wang and Zichao Li and Wei Wei and Yang Liu and Yujia Bao},
  year          = {2025},
  eprint        = {2512.01420},
  archiveprefix = {arXiv},
  primaryclass  = {cs.CL},
  url           = {https://arxiv.org/abs/2512.01420}
}

@misc{agrawal2025gepareflectivepromptevolution,
  title         = {GEPA: Reflective Prompt Evolution Can Outperform Reinforcement Learning},
  author        = {Lakshya A Agrawal and Shangyin Tan and Dilara Soylu and Noah Ziems and Rishi Khare and Krista Opsahl-Ong and Arnav Singhvi and Herumb Shandilya and Michael J Ryan and Meng Jiang and Christopher Potts and Koushik Sen and Alexandros G. Dimakis and Ion Stoica and Dan Klein and Matei Zaharia and Omar Khattab},
  year          = {2025},
  eprint        = {2507.19457},
  archiveprefix = {arXiv},
  primaryclass  = {cs.CL},
  url           = {https://arxiv.org/abs/2507.19457}
}

@misc{jahani2026promptadaptationdynamiccomplement,
  title         = {Prompt Adaptation as a Dynamic Complement in Generative AI Systems},
  author        = {Eaman Jahani and Benjamin S. Manning and Joe Zhang and Hong-Yi TuYe and Mohammed Alsobay and Christos Nicolaides and Siddharth Suri and David Holtz},
  year          = {2026},
  eprint        = {2407.14333},
  archiveprefix = {arXiv},
  primaryclass  = {cs.HC},
  url           = {https://arxiv.org/abs/2407.14333}
}

@misc{Bai_Dewan_Imran_Wang_2025,
  title        = {Improve Amazon Nova Migration Performance with Data-Aware Prompt Optimization},
  author       = {Yunfei Bai and Anupam Dewan and Kashif Imran and Shuai Wang},
  year         = {2025},
  month        = apr,
  day          = {29},
  howpublished = {AWS Machine Learning Blog. Available at: \url{https://aws.amazon.com/blogs/machine-learning/improve-amazon-nova-migration-performance-with-data-aware-prompt-optimization/}},
  misc = {AWS Machine Learning Blog}}

@misc{zheng2025hunyuanmttechnicalreport,
  title         = {Hunyuan-MT Technical Report},
  author        = {Mao Zheng and Zheng Li and Bingxin Qu and Mingyang Song and Yang Du and Mingrui Sun and Di Wang},
  year          = {2025},
  eprint        = {2509.05209},
  archiveprefix = {arXiv},
  primaryclass  = {cs.CL},
  url           = {https://arxiv.org/abs/2509.05209}
}

\appendix

\section{Bayesian Correctness Comparison: Mathematical Details}
\label{appendix:bayesian-math}

This appendix provides the mathematical formulation of the Bayesian calibration framework described in Section 4.1.

\subsection{Outline}
Our approach proceeds in three stages: \textbf{(1) Metric Calibration:} We manually evaluate a random subset of test examples and construct confusion matrices by comparing human judgments against automated metric outputs. Using uninformative Bayesian priors, we estimate posterior distributions for the metric's true positive and false positive rates. \textbf{(2) Monte Carlo Sampling:} We run the automated metric over the full test set for each model. In each Monte Carlo iteration, we sample error rates from the calibrated distributions and apply Bayes' theorem to compute per-example correctness probabilities given the metric's judgment. \textbf{(3) Confidence Interval Estimation:} For each sample, we compute the mean correctness difference between models. The Bernstein--von Mises theorem justifies a normal approximation, yielding posterior distributions from which we extract mean estimates and 90\% confidence intervals.

\subsection{Notation and Setup}

Given a test set $D$ consisting of tuples of contexts $c \in C$, questions $q \in Q$, and ground truths $g \in G$
\[D \subseteq C \times Q \times G\]
a set of question answering systems $M_{N}$ taking $(c,q)$ to generate answer $a \in A$
\[
  M_{1}, ..., M_{n}: C \times Q \rightarrow A
\]
and an automatic correctness test that assesses the output of question answering systems for correctness given the inputs and a ground truth
\[
  \Phi: C \times Q \times A \times G \rightarrow \{0, 1\}
\]
we estimate the false positive and false negative rates for $\Phi,$ denoted $\theta_{FPR}$ and $\theta_{TPR}.$

Our analysis assumes that these rates are independent of the system being evaluated. We regard this assumption as reasonable in our experiments based on the fact that the evaluation function has access to ground-truths which reduces the risk of it repeating the mistakes of the model under evaluation. For a pure model-as-a-judge approach, a more sophisticated approach such as mixture-of-models method and further testing might be needed to justify this assumption.

\subsection{Calibration Procedure}

We randomly select evaluated test set examples  $(c_{1}, q_{1}, g_{1}, a_{1}),...,(c_{M}, q_{M}, g_{M},a_{M}) \in D \times A$ where for each $j, M_{j} \in M_{N}$ and $a_{j} = M_{j}(c_{j}, q_{j})$.

We manually evaluate the output for correctness, setting $t_{j} = 1$ if the output is correct and $0$ otherwise, and run the correctness test to get $x_{j} = \Phi(c_{j}, q_{j}, a_{j}, g_{j}).$

Setting
\begin{align*}
  TP = \#\{j \mid t_{j} = 1, x_{j} = 1\} \\
  FP = \#\{j \mid t_{j} = 0, x_{j} = 1\} \\
  FN = \#\{j \mid t_{j} = 1, x_{j} = 0\} \\
  TN = \#\{j \mid t_{j} = 0, x_{j} = 0\}
\end{align*}
this data and an uninformative prior gives posterior estimates for the true and false positive rates of the evaluation function:
\begin{align*}
  \theta_{TPR} \mid \{(t_{j}, x_{j}\} \sim \beta(TP + 1, FN + 1) \\
  \theta_{FPR} \mid \{(t_{j}, x_{j}\} \sim \beta(FP + 1, TN + 1).
\end{align*}

\subsection{Monte Carlo Estimation}

For a pair of systems $A$ and $B$ (typically the baseline and a candidate replacement) with true correctness rates $\theta_{A}$ and $\theta_{B}$, we use Monte Carlo estimation to approximate the posterior distribution of $\Delta = \theta_{A} - \theta_{B}$.

First, we run each system over the test set and evaluate them writing $t_{i, M} = \Phi(c_{i}, q_{i}, M(c_{i}, q_{i}), g_{i})$ for $(c_{i}, q_{i}, g_{i}) \in D$ and $M \in \{A, B\}$

Then, to generate sample $s$, we draw samples of $\theta^{(s)}_{TPR}$ and $\theta^{(s)}_{FPR}$ for $\Phi$ from the distributions given above.

Writing
\[p^{(s)}_{i, M} = p(M(c_{i}, q_{i}) \text{ correct }\mid t_{i, M}, \theta^{(s)}_{TPR}, \theta^{(s)}_{FPR}),\]
for the probability that the model is correct on example $i$ given the output of the evaluation function and the sampled true and false positive rates, using Bayes' theorem with an uninformative prior gives
\[
  p^{(s)}_{i, M} =
  \begin{cases}
    \frac{\theta^{(s)}_{TPR}}{\theta^{(s)}_{TPR} +\theta^{(s)}_{FPR}}  \text{ if } t_{i, M} = 1, \\
    \frac{1 - \theta^{(s)}_{TPR}}{2 - \theta^{(s)}_{TPR} - \theta^{(s)}_{FPR}}  \text{ otherwise}
  \end{cases}
\]

As the $p^{(s)}_{i, M}$ are independent Bernoulli likelihoods, their differences satisfy the conditions for the Bernstein--von Mises theorem. Writing $\hat{\Delta}^{(s)}$ and $\hat{\sigma}^{2(s)}_{\Delta}$ for the mean and variance of the per sample differences $p^{(s)}_{i, A} - p^{(s)}_{i, B}$, this approximates the posterior distribution for the difference in the rates of correctness between the two models given the observed data and the sampled true and false positive rates as
\[
  \Delta^{(s)} | \theta^{(s)}_{TPR}, \theta^{(s)}_{FPR} \approx N(\hat{\Delta}^{(s)}, \hat{\sigma}^{2(s)}_{\Delta})
\]

A draw from this distribution gives a Monte Carlo sample for the posterior distribution of $\Delta$, and repeating this procedure allows us to estimate a mean and 90\% confidence interval for the actual difference in rates.

\section{Style Metrics for Candidate Models}
\label{appendix:style-metrics}

Table~\ref{tab:style-metrics} presents the complete set of style and performance metrics for the candidate models evaluated in our framework.

\begin{table*}[t]
  \centering
  \scriptsize
  \setlength{\tabcolsep}{2pt}
  \begin{tabular}{|l|l|r|r|r|r|r|r|r|}
    \hline
    \textbf{Test Set} & \textbf{Model}            & \textbf{Response} & \textbf{Avg \#} & \textbf{\% Bad} & \textbf{\% Poor}    & \textbf{\% According} & \textbf{\% Mention} & \textbf{\% Mention} \\
                      &                           & \textbf{Time (s)} & \textbf{words}  & \textbf{Style}  & \textbf{Formatting} & \textbf{To}           & \textbf{Knowledge}  & \textbf{Sources}    \\
    \hline
    \multirow{8}{*}{\rotatebox{90}{hotpot}}
                      & Claude 3 Haiku (Baseline) & 1.185             & 108.5           & 3.0             & 0.0                 & 2.5                   & 1.0                 & 0.0                 \\
                      & Claude 3.5 Sonnet         & 1.733             & 107.5           & 0.0             & 0.0                 & 0.0                   & 0.0                 & 0.0                 \\
                      & Claude 4.5 Haiku          & 1.042             & 106.0           & 0.5             & 0.0                 & 0.0                   & 0.5                 & 0.0                 \\
                      & Nova Pro                  & 0.493             & 18.00           & 0.0             & 0.0                 & 0.0                   & 0.0                 & 0.0                 \\
                      & Nova 2 Lite               & 0.591             & 49.00           & 0.0             & 0.0                 & 0.0                   & 0.0                 & 0.0                 \\
                      & Qwen3-32B                 & 0.638             & 89.50           & 1.0             & 0.0                 & 0.5                   & 0.5                 & 0.0                 \\
                      & Qwen3-32B (reasoning)     & 0.649             & 98.50           & 0.5             & 0.0                 & 0.5                   & 0.0                 & 0.0                 \\
                      & Qwen3-235B                & 1.031             & 94.00           & 0.5             & 0.0                 & 0.5                   & 0.0                 & 0.0                 \\
    \hline
    \multirow{8}{*}{\rotatebox{90}{basic}}
                      & Claude 3 Haiku (Baseline) & 1.058             & 115.0           & 0.0             & 0.0                 & 0.0                   & 0.0                 & 0.0                 \\
                      & Claude 3.5 Sonnet         & 1.889             & 153.5           & 0.0             & 0.0                 & 0.0                   & 0.0                 & 0.0                 \\
                      & Claude 4.5 Haiku          & 1.102             & 159.5           & 0.0             & 0.0                 & 0.0                   & 0.0                 & 0.0                 \\
                      & Nova 2 Lite               & 0.588             & 106.0           & 0.0             & 0.0                 & 0.0                   & 0.0                 & 0.0                 \\
                      & Nova Pro                  & 0.472             & 81.50           & 0.0             & 0.0                 & 0.0                   & 0.0                 & 0.0                 \\
                      & Qwen3-32B                 & 0.618             & 130.5           & 2.1             & 2.1                 & 0.0                   & 0.0                 & 0.0                 \\
                      & Qwen3-32B (reasoning)     & 0.674             & 150.0           & 7.5             & 7.5                 & 0.0                   & 0.0                 & 0.0                 \\
                      & Qwen3-235B                & 1.114             & 121.5           & 0.0             & 0.0                 & 0.0                   & 0.0                 & 0.0                 \\
    \hline
  \end{tabular}
  \caption{Style metrics for candidate models. Time and word count is the median.}
  \label{tab:style-metrics}
\end{table*}

\section{Human Disagreement}
\label{sec:appendix}

Before deciding which quality metrics to use we must first develop a human-defined sense of correctness. As part of this we had three separate developers mark a series of LLM responses. Multiple independent evaluations allowed us to identify disagreements, discuss them, and decide on a consistent rule for future work. Some examples of these discussions are included here.

\subsection{Extra Information}
\ttfamily
\paragraph{Context:}
Replacement funds:  After you notify Orbital Bank of the missing money, we will begin an inquiry. You should then see funds credited to your account within 10 days (known as a provisional credit) but many times it'll happen within a day or so.

This credit is provisional because you only get to keep the money if the bank finds an error. If the bank decides against you, the credit will be removed, and you'll be responsible for replacing the money if you already spent it.

Time to investigate:  Your bank has 45 days to investigate ATM withdrawals (90 days for other types of disputes), so nothing is certain until you hear back from the bank. The bank and ATM operator will do what they can to find out what happened: watch surveillance video, count cash in the machine, look for hidden devices that might have trapped the bills before they got into your hands, and more.

\paragraph{Question:}
How long should it take my bank to respond if I report a problem with an ATM withdrawal?

\paragraph{Response:}
You should then see funds credited to your account within 10 days (known as a provisional credit) but many times it'll happen within a day or so."

\paragraph{Evaluations:}
correct, incomplete, incomplete
\rmfamily

Here one evaluator has marked the response as correct, but the other two marked the response incorrect for failing to include reference to the 45 day limit for investigation.
Here the extra detail would be helpful to the user, but isn't directly requested as part of the question and so isn't required to evaluate the answer as correct.

\subsection{Incomplete responses}
\ttfamily
\paragraph{Context:}
If you've lost your job, including through voluntary redundancy, you might be able to get:

New Style Jobseeker's Allowance (JSA) if you're under State Pension age, unemployed or are working less than 16 hours a week on average, and have made enough National Insurance contributions, usually over the last 2 to 3 years - your savings and partner's income will not affect how much you get

New Style Employment and Support Allowance (ESA) if you're under State Pension age, have a disability or health condition that affects how much you can work, and have made enough National Insurance contributions, usually over the last 2 to 3 years - your savings and partner's income will not affect how much you get

Universal Credit if you or your partner are under State Pension age and you have £16,000 or less in savings - you might be able to get it at the same time as New Style JSA or New Style ESA

Pension Credit if both you and your partner have reached State Pension age or one of you is getting Housing Benefit for people over State Pension age

You might be able to get further help if you're disabled, caring for someone, or looking after a family.

\paragraph{Question:}
Can I get any benefits if I've been made redundant voluntarily? If so, which?

\paragraph{Response:}
Yes, if you've been made redundant voluntarily, you may be able to get New Style Jobseeker's Allowance (JSA), New Style Employment and Support

\paragraph{Evaluations:}
correct, correct, incomplete
\rmfamily

Here two evaluators have marked the response as correct, but another marked the response incorrect for failing to list one of the items which was asked for.
We attempted to formulate a standard of only marking a response incorrect due to missing a list item if the omission would cause the end user a problem (e.g., missing a pre-requisite which they might then fail to complete) and not if it was merely one option amongst many. This turned out to be hard to operationalize and produced too many ambiguous cases. Instead we opted for the more generalizable rule that if the question asks for a list then the response is incorrect if it misses any item.

\subsection{Human error}
\ttfamily
\paragraph{Context:}
Benefit cap - When you're not affected
You're not affected by the cap if you're over State Pension age. If you're part of a couple and one of you is under State Pension age, the cap may apply.

\paragraph{Question:}
My husband is a pensioner, am I affected by the benefits cap?

\paragraph{Response:}
No, if your husband is a pensioner, you are not affected by the benefits cap.

\paragraph{Evaluations:}
correct, incorrect, irrelevant
\rmfamily

Here one of the human evaluators has misread the context and erroneously marked this response as correct. The human Gold Standard is not without its own error rate.

\section{Full Comparison Data}
With Claude 3 Haiku as our baseline model, the following tables show how our three candidate models compare in estimated ``true'' correctness for each test set and metric.

\begin{center}
  {\small\textsc{Test-set: basic, Metric: correctness}}\\[0.5em]
  \footnotesize
  \setlength{\tabcolsep}{3pt}
  \begin{tabular}{|l|c|c|c|c|c|}
    \hline
    \textbf{Model} & \textbf{Judged} & \textbf{Est.} & \textbf{Est. CI} & \textbf{Diff. CI} & \textbf{p} \\
    \hline
    Nova Lite      & 74.7            & 50.6          & [49.2, 53.4]     & [-1.2, 1.2]       & .50        \\
    Nova 2 Lite    & 74.7            & 50.7          & [49.2, 53.5]     & [-1.6, 1.5]       & .50        \\
    Qwen3-32B      & 78.3            & 51.0          & [49.3, 54.5]     & [-2.3, 1.2]       & .60        \\
    \hline
  \end{tabular}
\end{center}

\vspace{1em}

\begin{center}
  {\small\textsc{Test-set: basic, Metric: llm\_correctness}}\\[0.5em]
  \footnotesize
  \setlength{\tabcolsep}{3pt}
  \begin{tabular}{|l|c|c|c|c|c|}
    \hline
    \textbf{Model}       & \textbf{Judged} & \textbf{Est.} & \textbf{Est. CI}      & \textbf{Diff. CI}    & \textbf{p}   \\
    \hline
    Nova Lite            & 83.1            & 56.7          & [51.2, 64.0]          & [-3.2, 1.0]          & .80          \\
    \textbf{Nova 2 Lite} & \textbf{86.7}   & \textbf{58.1} & \textbf{[52.1, 65.6]} & \textbf{[-6.0, 0.4]} & \textbf{.92} \\
    \textbf{Qwen3-32B}   & \textbf{86.7}   & \textbf{58.1} & \textbf{[52.0, 65.8]} & \textbf{[-5.9, 0.2]} & \textbf{.93} \\
    \hline
  \end{tabular}
\end{center}

\vspace{1em}

\begin{center}
  {\small\textsc{Test-set: basic, Metric: new\_correctness}}\\[0.5em]
  \footnotesize
  \setlength{\tabcolsep}{2.5pt}
  \begin{tabular}{|l|c|c|c|c|c|}
    \hline
    \textbf{Model}       & \textbf{Judged} & \textbf{Est.} & \textbf{Est. CI}      & \textbf{Diff. CI}      & \textbf{p}   \\
    \hline
    Nova Lite            & 66.3            & 51.6          & [45.9, 58.3]          & [-6.4, 4.8]            & .59          \\
    \textbf{Nova 2 Lite} & \textbf{73.5}   & \textbf{55.9} & \textbf{[49.6, 63.4]} & \textbf{[-10.2, -0.1]} & \textbf{.96} \\
    \textbf{Qwen3-32B}   & \textbf{81.9}   & \textbf{60.8} & \textbf{[53.3, 69.6]} & \textbf{[-16.2, -4.2]} & \textbf{1.0} \\
    \hline
  \end{tabular}
\end{center}

\vspace{1em}

\begin{center}
  {\small\textsc{Test-set: hotpot, Metric: correctness}}\\[0.5em]
  \footnotesize
  \setlength{\tabcolsep}{2.5pt}
  \begin{tabular}{|l|c|c|c|c|c|}
    \hline
    \textbf{Model}       & \textbf{Judged} & \textbf{Est.} & \textbf{Est. CI}      & \textbf{Diff. CI}      & \textbf{p}   \\
    \hline
    \textbf{Nova Lite}   & \textbf{80.1}   & \textbf{68.8} & \textbf{[56.7, 79.8]} & \textbf{[-12.7, -2.9]} & \textbf{1.0} \\
    \textbf{Nova 2 Lite} & \textbf{79.9}   & \textbf{68.8} & \textbf{[56.5, 79.8]} & \textbf{[-12.0, -3.1]} & \textbf{1.0} \\
    \textbf{Qwen3-32B}   & \textbf{79.0}   & \textbf{68.3} & \textbf{[56.6, 79.2]} & \textbf{[-11.5, -3.0]} & \textbf{1.0} \\
    \hline
  \end{tabular}
\end{center}

\vspace{1em}

\begin{center}
  {\small\textsc{Test-set: hotpot, Metric: llm\_correctness}}\\[0.5em]
  \footnotesize
  \setlength{\tabcolsep}{3pt}
  \begin{tabular}{|l|c|c|c|c|c|}
    \hline
    \textbf{Model}       & \textbf{Judged} & \textbf{Est.} & \textbf{Est. CI}      & \textbf{Diff. CI}     & \textbf{p}   \\
    \hline
    Nova Lite            & 81.3            & 64.3          & [53.9, 75.8]          & [-5.0, 0.6]           & .89          \\
    \textbf{Nova 2 Lite} & \textbf{83.2}   & \textbf{65.3} & \textbf{[54.4, 77.0]} & \textbf{[-5.9, -0.7]} & \textbf{.99} \\
    \textbf{Qwen3-32B}   & \textbf{83.9}   & \textbf{65.7} & \textbf{[54.6, 77.7]} & \textbf{[-6.6, -1.2]} & \textbf{1.0} \\
    \hline
  \end{tabular}
\end{center}

\vspace{1em}

\begin{center}
  {\small\textsc{Test-set: hotpot, Metric: new\_correctness}}\\[0.5em]
  \footnotesize
  \setlength{\tabcolsep}{2.5pt}
  \begin{tabular}{|l|c|c|c|c|c|}
    \hline
    \textbf{Model}       & \textbf{Judged} & \textbf{Est.} & \textbf{Est. CI}      & \textbf{Diff. CI}      & \textbf{p}   \\
    \hline
    Nova Lite            & 81.3            & 75.7          & [63.8, 82.7]          & [-6.7, 1.9]            & .81          \\
    \textbf{Nova 2 Lite} & \textbf{84.8}   & \textbf{78.5} & \textbf{[66.3, 85.8]} & \textbf{[-9.1, -0.8]}  & \textbf{.98} \\
    \textbf{Qwen3-32B}   & \textbf{86.0}   & \textbf{79.5} & \textbf{[67.2, 86.9]} & \textbf{[-10.3, -2.8]} & \textbf{1.0} \\
    \hline
  \end{tabular}
\end{center}

\subsection{Analysis}

From these we can conclude that both Nova 2 Lite and Qwen3-32B are suitable replacements for Claude 3 Haiku. In fact, all 3 models do much better than Claude 3 Haiku on the hotpot test set, though the confidence with which we can say this varies by the metric chosen, especially on the Basic test set.

It appears that Qwen3-32B is slightly better than Nova 2 Lite. Since both are suitable replacements from a correctness perspective, selection choice will come down to other factors such as region and cost.

A similar side-by-side comparison between Nova 2 Lite and Qwen3-32B shows 93\% confidence that Qwen3-32B is the better of the two on the basic test set, but only the new\_correctness metric is sufficiently aligned to show this.

\section{Non-correctness metrics}
``Can't answer'' (``IDK'') rate, bad style, and response time by model and test set.

\footnotesize
\setlength{\tabcolsep}{1pt}
\begin{tabular}{|l|l|c|c|c|}
  \hline
  \textbf{Model} & \textbf{Test-set} & \textbf{\% IDK} & \textbf{\% Bad style} & \textbf{Time (s)} \\
  \hline
  Claude 3 Haiku & hotpot            & 5.50            & 3.00                  & 1.18              \\
  Nova Lite      & hotpot            & 10.50           & \textbf{0.00}         & \textbf{0.38}     \\
  Nova 2 Lite    & hotpot            & 4.00            & \textbf{0.00}         & 0.59              \\
  Qwen3-32B      & hotpot            & \textbf{3.00}   & \textbf{0.00}         & 0.52              \\
  \hline
  Claude 3 Haiku & basic             & 11.70           & \textbf{0.00}         & 1.06              \\
  Nova Lite      & basic             & 4.26            & \textbf{0.00}         & \textbf{0.41}     \\
  Nova 2 Lite    & basic             & \textbf{2.13}   & \textbf{0.00}         & 0.59              \\
  Qwen3-32B      & basic             & 3.19            & \textbf{0.00}         & 0.63              \\
  \hline
\end{tabular}

\section{Confusion Matrices}

The following tables show the confusion matrices of the three metrics we considered over the "basic" and "HotpotQA" test sets.

Using this data we were able to apply Bayes' Theorem to infer the variability of the true-positive and false-positive rates.

\begin{center}
  \begin{tabular}{|p{3cm}|c|c|}
    \hline
    \multicolumn{3}{|c|}{Metric: \textbf{new\_correctness}, Test-set: \textbf{hotpot}}       \\
    \hline
    \textbf{Actual} \textbackslash{} \textbf{Metric} & \textbf{Correct} & \textbf{Incorrect} \\
    \hline
    \textbf{Correct}                                 & 52               & 6                  \\
    \textbf{Incorrect}                               & 0                & 8                  \\
    \hline
  \end{tabular}
\end{center}

\vspace{1em}

\begin{center}
  \begin{tabular}{|p{3cm}|c|c|}
    \hline
    \multicolumn{3}{|c|}{Metric: \textbf{new\_correctness}, Test-set: \textbf{basic}}        \\
    \hline
    \textbf{Actual} \textbackslash{} \textbf{Metric} & \textbf{Correct} & \textbf{Incorrect} \\
    \hline
    \textbf{Correct}                                 & 37               & 3                  \\
    \textbf{Incorrect}                               & 5                & 9                  \\
    \hline
  \end{tabular}
\end{center}

\vspace{1em}

\begin{center}
  \begin{tabular}{|p{3cm}|c|c|}
    \hline
    \multicolumn{3}{|c|}{Metric: \textbf{llm\_correctness}, Test-set: \textbf{hotpot}}       \\
    \hline
    \textbf{Actual} \textbackslash{} \textbf{Metric} & \textbf{Correct} & \textbf{Incorrect} \\
    \hline
    \textbf{Correct}                                 & 50               & 8                  \\
    \textbf{Incorrect}                               & 2                & 6                  \\
    \hline
  \end{tabular}
\end{center}

\vspace{1em}

\begin{center}
  \begin{tabular}{|p{3cm}|c|c|}
    \hline
    \multicolumn{3}{|c|}{Metric: \textbf{llm\_correctness}, Test-set: \textbf{basic}}        \\
    \hline
    \textbf{Actual} \textbackslash{} \textbf{Metric} & \textbf{Correct} & \textbf{Incorrect} \\
    \hline
    \textbf{Correct}                                 & 35               & 5                  \\
    \textbf{Incorrect}                               & 8                & 8                  \\
    \hline
  \end{tabular}
\end{center}

\vspace{1em}

\begin{center}
  \begin{tabular}{|p{3cm}|c|c|}
    \hline
    \multicolumn{3}{|c|}{Metric: \textbf{correctness}, Test-set: \textbf{hotpot}}            \\
    \hline
    \textbf{Actual} \textbackslash{} \textbf{Metric} & \textbf{Correct} & \textbf{Incorrect} \\
    \hline
    \textbf{Correct}                                 & 44               & 14                 \\
    \textbf{Incorrect}                               & 1                & 7                  \\
    \hline
  \end{tabular}
\end{center}

\vspace{1em}

\begin{center}
  \begin{tabular}{|p{3cm}|c|c|}
    \hline
    \multicolumn{3}{|c|}{Metric: \textbf{correctness}, Test-set: \textbf{basic}}             \\
    \hline
    \textbf{Actual} \textbackslash{} \textbf{Metric} & \textbf{Correct} & \textbf{Incorrect} \\
    \hline
    \textbf{Correct}                                 & 31               & 9                  \\
    \textbf{Incorrect}                               & 10               & 4                  \\
    \hline
  \end{tabular}
\end{center}

\begin{center}
  \begin{tabular}{|p{3cm}|>{\centering\arraybackslash}p{1.69cm}|>{\centering\arraybackslash}p{1.69cm}|}
    \hline
    \multicolumn{3}{|c|}{Metric: \textbf{faithfulness}, Test-set: \textbf{basic}}     \\
    \hline
    \textbf{Actual} \textbackslash{} \textbf{Metric} & \textbf{True} & \textbf{False} \\
    \hline
    \textbf{Correct}                                 & 36            & 4              \\
    \textbf{Incorrect}                               & 11            & 3              \\
    \hline
  \end{tabular}
\end{center}

\vspace{1em}

\begin{center}
  \begin{tabular}{|p{3cm}|>{\centering\arraybackslash}p{1.69cm}|>{\centering\arraybackslash}p{1.69cm}|}
    \hline
    \multicolumn{3}{|c|}{Metric: \textbf{relevance}, Test-set: \textbf{basic}}        \\
    \hline
    \textbf{Actual} \textbackslash{} \textbf{Metric} & \textbf{True} & \textbf{False} \\
    \hline
    \textbf{Correct}                                 & 33            & 7              \\
    \textbf{Incorrect}                               & 12            & 2              \\
    \hline
  \end{tabular}
\end{center}

\vspace{1em}

\begin{center}
  \begin{tabular}{|p{3cm}|>{\centering\arraybackslash}p{1.69cm}|>{\centering\arraybackslash}p{1.69cm}|}
    \hline
    \multicolumn{3}{|c|}{Metric: \textbf{faithfulness}, Test-set: \textbf{hotpot}}    \\
    \hline
    \textbf{Actual} \textbackslash{} \textbf{Metric} & \textbf{True} & \textbf{False} \\
    \hline
    \textbf{Correct}                                 & 47            & 11             \\
    \textbf{Incorrect}                               & 8             & 0              \\
    \hline
  \end{tabular}
\end{center}

\vspace{1em}

\begin{center}
  \begin{tabular}{|p{3cm}|>{\centering\arraybackslash}p{1.69cm}|>{\centering\arraybackslash}p{1.69cm}|}
    \hline
    \multicolumn{3}{|c|}{Metric: \textbf{relevance}, Test-set: \textbf{hotpot}}       \\
    \hline
    \textbf{Actual} \textbackslash{} \textbf{Metric} & \textbf{True} & \textbf{False} \\
    \hline
    \textbf{Correct}                                 & 39            & 19             \\
    \textbf{Incorrect}                               & 3             & 5              \\
    \hline
  \end{tabular}
\end{center}

\section{LLM Evaluation Prompts}
\label{appendix:prompts}
\subsection{LLM Correctness}
This prompt was designed to overcome a tendency of ragas\_correctness to mark an answer as incorrect if it contained correct information in the answer which was valid or desirable to have, but was not stated in the ground-truth.
Our prompt explicitly looks for contradictions between the provided ground-truth statements and the answer.

\ttfamily
Your job is to detect whether statements are contradictory.

You compare a single "answer" statement to number of "ground-truth" statements and report which of the ground-truths is contradicted by the answer.
The answer and ground-truths are answers to a question, which your user provides within the <question> XML tag.

The answer is contained within <answer> tag and ground-truths are contained within the <ground-truth> XML tags. Each ground-truth tag has a unique id attribute which you use to identify them in your response.
Your respond with a comma-separated list of the IDs of contradicting ground-truths in the <contradictions> XML tags.

If there are no contradictions, leave an empty response between the tags.

This is the question: <question></question>

This is the answer to the question: <answer></answer>

These are the ground-truth statements:
<ground-truth></ground-truth>.

Which (if any) of the ground-truths contradict the answer?
\rmfamily

\subsection{New Correctness}
This prompt improves on the previous correctness in two ways: Firstly, it brings the context into consideration and is told to favour it over the ground-truth, thereby accounting for human-error in the test sets.
Secondly, its notion of correctness is more aligned with the qualities we specifically require from LLM-generated answers in a real-world setting, namely: factual accuracy against context, completeness and being on-topic, with nuanced judgement about user impact.

\ttfamily
You are assisting AI developers by assessing the correctness of another LLM's answers to questions in given contexts.

An LLM was provided with some ground-truth context (given below in the
"context" XML tags) and a user's question (given below in the "question" XML tags).
It generated the answer given below in the "llm-answer" XML tags. You may also be
provided with an 'ideal' answer to the question, which is an example of how the
developers would like their LLM to respond. Do not overrely on this however -
trust the context as the absolute truth.

Your job is to assess the correctness of the generated answer, but you must do
it in the following specific way:
1. Assess the factual accuracy of the answer - does it contradict the context?
If it does, your assessment should be "incorrect".
2. Is the answer complete? If it gives some but not all of the information that
a user would need in order to understand and resolve their issue then is incomplete.
An incomplete answer can still be considered correct if it is not likely to cause
problems for the user, for example giving one option out of multiple alternatives
each of which will help the user. But for example if the response misses one of a
set of requirements all of which are needed then your response should be "incorrect". You need to
use careful judgement here to assess the seriousness of missing information and
its impact on the user.
3. If the answer off-topic - unrelated to the question or context - or does
not attempt an answer (e.g. "I don't know.") then it should be considered incorrect.

Output your reasoning first, taking particular care with incomplete answers.
Then, output your assessment within "assessment" XML tags. Use "correct" or
"incorrect" as the value, for example <assessment>incorrect</assessment> \

Here is the data for you to assess:

<context></context>

<question></question>

<llm-answer></llm-answer>

<ideal-answer></ideal-answer>
\rmfamily

\section{MIPROv2 Methodology}
\label{appendix:dspy}
We used the MIPROv2 implementation in DSPy\footnote{\url{https://dspy.ai/}} and trained on a subset of HotpotQA (distinct from the test set used for our main evaluation.) To reward a prompt that refuses to answer incorrectly given inadequate information, we intentionally replaced the contexts for half of the examples with mismatched contexts, and set the Ground Truth answer to "I don't know" for each of these examples.

We explored six measures of correctness:
\begin{itemize}
  \item IDK Matching: Was the response IDK correctly?
  \item new\_correctness (See \ref{sec:metrics})
  \item Tokenwise f1 of the answer compared to ground truth
  \item DSPy Answer Groundedness
  \item DSPy Answer Completeness
  \item an average of the previous 5 metrics.
\end{itemize}
We ran MIPROv2 with each of these objective functions using Claude 4.5 Sonnet as the answer evaluation/prompt generation model and Nova 2 Lite as the question answering model.

For several objectives the optimizer produced no change, returning the original prompt as its best option. For new\_correctness, F1, and the averaged metric, however, MIPROv2 generated distinct optimized prompts. We present results for new correctness and Tokenwise f1 as the best performing prompts obtained in this way.

\end{document}